\def\paperlanguage{}
\pdfoutput=1 % for arxiv

\documentclass[letterpaper, 10 pt, conference]{ieeeconf}  % Comment this line out if you need a4paper

\usepackage{bm}
\usepackage{cite}
\include{preamble}

\IEEEoverridecommandlockouts                              % This command is only needed if
\title{\LARGE \textbf
  {
    \switchlanguage%
    {%
      Learning-Based Wiping Behavior of Low-Rigidity Robots Considering Various Surface Materials and Task Definitions
    }%
    {%
      様々な表面とタスク定義を考慮した\\低剛性ロボットの学習型拭き行動の実現
    }%
  }
}

\author{Kento Kawaharazuka$^{1}$, Naoaki Kanazawa$^{1}$, Kei Okada$^{1}$, and Masayuki Inaba$^{1}$% <-this % stops a space
  \thanks{$^{1}$ The authors are with the Department of Mechano-Informatics, Graduate School of Information Science and Technology, The University of Tokyo, 7-3-1 Hongo, Bunkyo-ku, Tokyo, 113-8656, Japan.
    {\texttt\small [kawaharazuka, kanazawa, k-okada, inaba]@jsk.t.u-tokyo.ac.jp}
  }
}
\begin{document}

\maketitle
\thispagestyle{empty}
\pagestyle{empty}

%%%%%%%%%%%%%%%%%%%%%%%%%%%%%%%%%%%%%%%%%%%%%%%%%%%%%%%%%%%%%%%%%%%%%%%%%%%%%%%%
\begin{abstract}
  \switchlanguage%
  {%
    Wiping behavior is a task of tracing the surface of an object while feeling the force with the palm of the hand.
    It is necessary to adjust the force and posture appropriately considering the various contact conditions felt by the hand.
    Several studies have been conducted on the wiping motion, however, these studies have only dealt with a single surface material, and have only considered the application of the amount of appropriate force, lacking intelligent movements to ensure that the force is applied either evenly to the entire surface or to a certain area.
    Depending on the surface material, the hand posture and pressing force should be varied appropriately, and this is highly dependent on the definition of the task.
    Also, most of the movements are executed by high-rigidity robots that are easy to model, and few movements are executed by robots that are low-rigidity but therefore have a small risk of damage due to excessive contact.
    So, in this study, we develop a method of motion generation based on the learned prediction of contact force during the wiping motion of a low-rigidity robot.
    We show that MyCobot, which is made of low-rigidity resin, can appropriately perform wiping behaviors on a plane with multiple surface materials based on various task definitions.
  }%
  {%
    拭き動作は手の平で力を感じながら物体表面をなぞるような動作である.
    手に感じる多様な接触状態を考慮し適切に力や角度を調整する必要がある.
    これまで拭き動作についていくつかの研究が行われてきた.
    一方で, それらはある単一の表面素材のみを扱っており, また, 適切な力をかけるばかりで表面全体に満遍なく力がかかっているかやそれを偏らせるような知能的な動作に乏しい.
    表面素材によっても手の角度や押しつけ力は適切に変化させるべきであるし, これはそのタスクの定義にも大きく依存する.
    そして, モデル化容易な高剛性ロボットによる動作実行がほとんどであり, 低剛性だがそれゆえに過度な接触による破損のリスクも小さなロボットによる実行は少ない.
    そこで本研究では, 低剛性なロボットの拭き動作における接触力変化を学習し, これをもとに動作生成を行う手法を開発する.
    低剛性樹脂製のMyCobotが, 複数の表面素材を持つ平面に対して, 様々なタスク定義から適切に拭き行動を実現できることを実機実験により示す.
  }%
\end{abstract}

\section{INTRODUCTION}\label{sec:introduction}
\switchlanguage%
{%
  Wiping behavior is a basic human motion of tracing the surface of an object while feeling the force with the palm of the hand.
  It is necessary to adjust the force, hand posture, etc. appropriately considering the various contact conditions felt by the hand.
  Several studies have been conducted on this wiping motion \cite{kim2019cleaning}.
  \cite{hess2012coverage} has formulated and discussed the generation of trajectories for wiping 3D surfaces as a traveling salesman problem.
  \cite{domentios2018wiping} has developed a method for generating trajectories that wipe 3D surfaces using dynamic motion primitives and avoiding restricted areas.
  \cite{saito2020wiping} has generated wiping motions on 2D surfaces using imitation learning based on visual, tactile, and joint information, and extended this method to 3D surfaces including occlusion \cite{saito2022wiping}.
  \cite{martin2019variable} has generated wiping motions on a 2D plane by reinforcement learning with variable impedance control as a control input.
  \cite{wang2021variable} furthermore automatically adjusts the gain of the variable impedance control to enable wiping operation on several surface materials.
  As shown above, it is now possible to generate trajectories using a general geometric model, position control, torque control, imitation learning, and reinforcement learning.

  On the other hand, we believe that these studies have three problems.
  (1) They usually deal with only a single type of surface material and lack flexibility.
  In reality, the surfaces to be wiped vary from smooth floors to rough desks and uneven walls.
  (2) They only apply a constant force in an appropriate direction, and lack intelligent motion generation such as checking whether the force is applied either evenly to the entire contact surface, or daring to a certain area.
  (3) Most of the experiments have been conducted with high-rigidity robots that are easy to model and can execute movements accurately, but not with robots that are less rigid and therefore have less risk of damage due to excessive contact.
  If this becomes possible, the task can be performed with less expensive and less capable robots.
  The objective of this study is to solve these problems (1)--(3) to realize a more adaptive wiping motion.
}%
{%
  拭き動作は手の平で力を感じながら物体表面をなぞる人間の基本的な動作である.
  手に感じる多様な接触状態を考慮し, 適切に力や手の角度等を調整する必要がある.
  これまで, この拭き動作についていくつかの研究が行われてきた\cite{kim2019cleaning}.
  \cite{hess2012coverage}は3次元曲面を拭く軌道の生成を巡回セールスマン問題として定式化し議論している.
  \cite{domentios2018wiping}は3次元曲面を拭く軌道の生成をdynamic motion primitivesを用いて行い, restricted areasを避けながら動作する手法を開発している.
  \cite{saito2020wiping}は2次元曲面を拭く動作を視覚・触覚・関節情報から模倣学習を用いて生成し, これをオクルージョンを含む3次元曲面へと発展させている\cite{saito2022wiping}.
  \cite{martin2019variable}は2次元平面を拭く動作をvariable impedance制御を入力とした強化学習により生成している.
  \cite{wang2021variable}ではさらにvariable impedance controlのゲインを自動で調整することで多様な表面素材の拭き動作を可能にしている.
  これらのように, 一般的な幾何モデルを用いた軌道生成と位置制御やトルク制御, 模倣学習や強化学習を用いた拭き動作の実現が可能となってきた.

  一方で, これらの研究には3つの問題点(1)--(3)があると考えている.
  (1) ある単一の表面素材のみを扱っており, その柔軟性に欠ける.
  拭き動作を行う表面は, ツルツルの床からザラザラの机, 凸凹の壁まで様々である.
  (2) 適切な方向に一定の力をかけるばかりで, 接触表面全体に満遍なく力がかかっているかを確認したり, 敢えてそれを偏らせたりするような知能的な動作生成に乏しい.
  (3) モデル化容易で正確に動作実行可能な高剛性ロボットによる実験がほとんどであり, 低剛性だがそれゆえに過度な接触による破損リスクも小さいロボットによる実行も行われていない.
  これが可能になれば, より安価で性能の悪いロボットでもタスク実行が可能となる.
  これら(1)--(3)の問題点を解決することで, より適応的な拭き動作を実現することを本研究の目的とする.
}%

\begin{figure}[t]
  \centering
  \includegraphics[width=0.95\columnwidth]{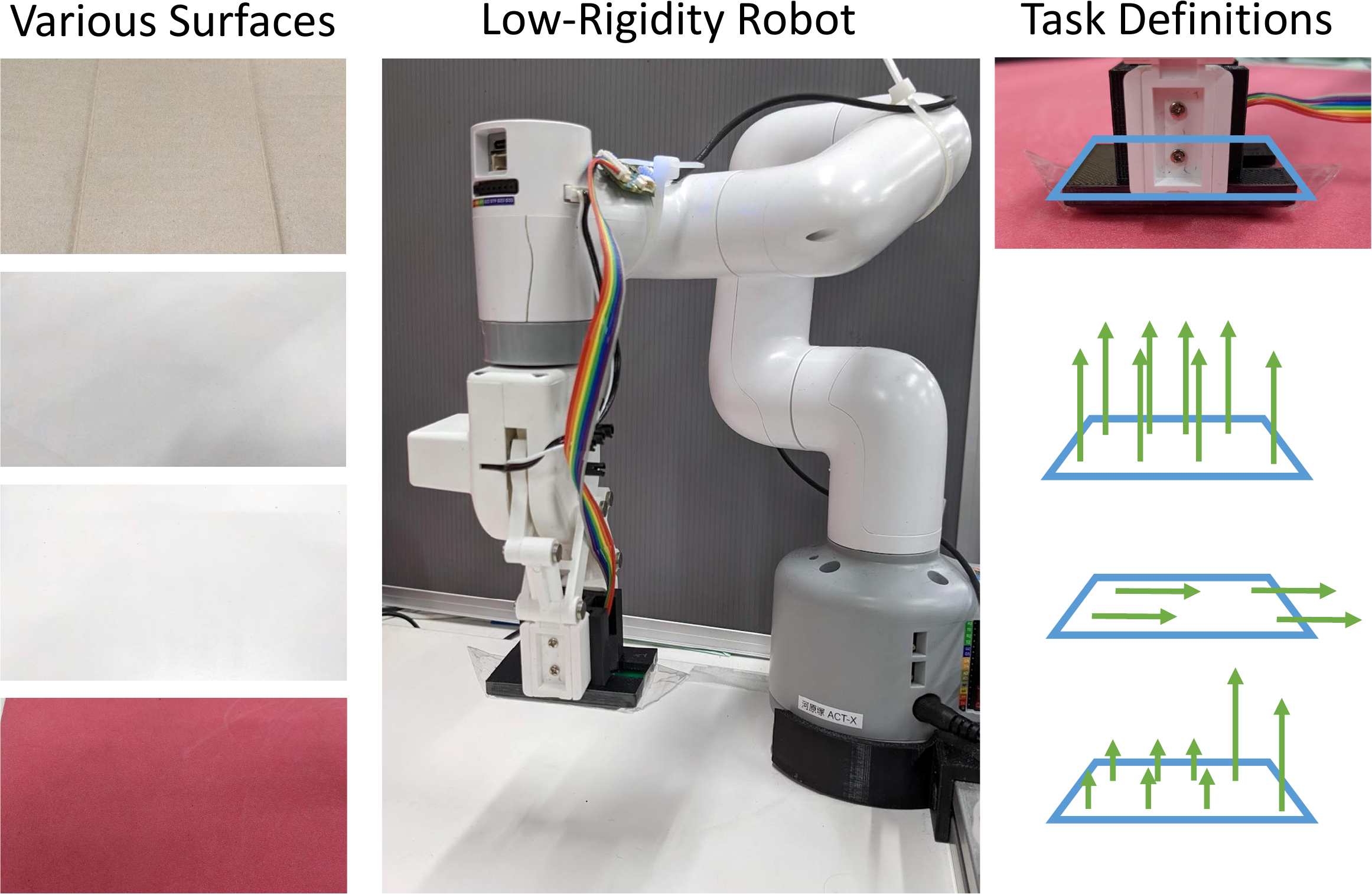}
  \caption{The concept of this study: we handle wiping behaviors with (1) various surface materials, (2) various task definitions, and (3) low-rigidity robot.}
  \label{figure:concept}
  \vspace{-1.0ex}
\end{figure}

\begin{figure*}[t]
  \centering
  \includegraphics[width=1.95\columnwidth]{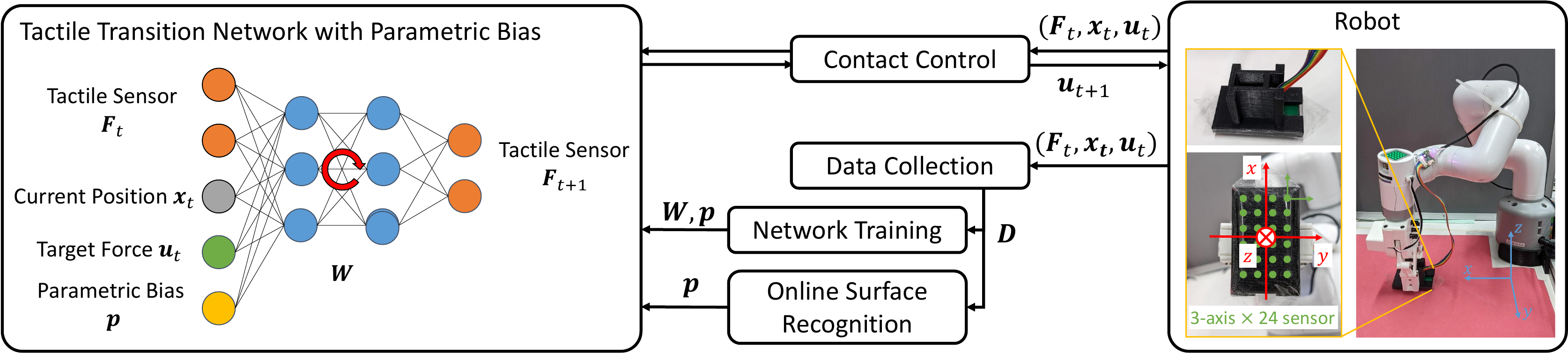}
  \caption{The overview of the learning system of this study.}
  \label{figure:system}
  \vspace{-1.0ex}
\end{figure*}

\switchlanguage%
{%
  To solve (3), we conduct experiments using an actual robot, MyCobot, which is made of low-rigidity resin and has difficulty in performing accurate movements.
  We collect data from the motion of the robot by adding random noise to a simple proportional control, and construct a contact transition model by learning a neural network, which takes into account the softness and rattling of the robot.
  To solve (1), we use a learnable input variable, parametric bias \cite{tani2002parametric, tani2004parametric}, which can extract multiple attractor dynamics from various data, to deal with different surface materials.
  Parametric bias has been mainly used for imitation learning, but we are now applying it to predictive model learning \cite{kawaharazuka2020dynamics, kawaharazuka2022cloth, kawaharazuka2022vservoing}.
  To solve (2), we use a contact sensor, uSkin \cite{tomo2018uskin}, which contains 24 3-axis sensors in an area of 31 mm $\times$ 51.5 mm.
  We show that various task definitions can be created for the contact and friction forces by setting appropriate loss functions for these 72 dimensional values.
  In order to focus on solving (1)--(3), we limit our experiments to a two-dimensional plane instead of a three-dimensional surface.

  This study is organized as follows.
  In \secref{sec:introduction}, the background and purpose of this study are described.
  In \secref{sec:proposed}, we describe a network configuration for representing the state transitions of contact sensors, data collection with random noise added to proportional control, network training, recognition of surface material properties, and contact control for various task definitions.
  In \secref{sec:experiment}, we conduct experiments using an actual low-rigidity robot, MyCobot.
  Finally, in \secref{sec:discussion}, we discuss the results, and in \secref{sec:conclusion}, we conclude and give some future perspectives.
}%
{%
  (3)について, 本研究では低剛性樹脂製で正確な動作の難しいロボットMyCobotの実機を用いて実験を行う.
  簡易な比例制御にランダムノイズを加えた動きからデータを収集し, ニューラルネットワークの学習によって接触遷移モデルを構築することで, ロボットの柔らかさやガタ等まで考慮に入れた学習を行う.
  (1)について, 様々なデータから複数のアトラクターダイナミクスを抽出可能な入力変数Parametric Bias \cite{tani2002parametric}を用いることで, 様々な表面素材に対応する.
  Parametric Biasについてはこれまで模倣学習への応用が主であったが, 現在予測モデル学習への適用を進めており, これらを応用する\cite{kawaharazuka2020dynamics, kawaharazuka2022cloth, kawaharazuka2022vservoing}.
  (2)について, 本研究では接触センサとしてたった31 mm $\times$ 51.5 mmの中に3軸センサが24個入ったuSkin \cite{tomo2018uskin}を用いる.
  この72次元の値に対して, 適切に損失関数を設定することで, 接触力や摩擦力, その偏りに関して様々なタスク定義を行うことができることを示す.
  なお, 本研究では(1)--(3)に焦点を当てるため, 3次元曲面ではなく2次元平面に限定して実験を行う.

  本研究は以下のような構成となっている。
  \secref{sec:introduction}では本論文の背景と目的について述べた.
  \secref{sec:proposed}では接触センサの状態遷移を表現するネットワーク構成, 比例制御にランダムノイズを加えたデータ収集, ネットワーク訓練, 表面素材特性の推定, 様々な目的に応じた接触制御について順に述べる.
  \secref{sec:experiment}では低剛性ロボットMyCobotを使った実機動作実験を行う.
  最後に、\secref{sec:discussion}で考察を述べ, \secref{sec:conclusion}で結論と今後の展望を述べる.
}%

\section{Learning-Based Wiping Behavior of Low-Rigidity Robots Considering Various Surfaces and Task Definitions} \label{sec:proposed}
\switchlanguage%
{%
  The overall system of this study is shown in \figref{figure:system}.
  In this study, a low-rigidity robot MyCobot is used.
  As shown in the right figure of \figref{figure:system}, MyCobot is equipped with a contact sensor, uSkin, which contains 24 3-axis force sensors on its tip end, which is pressed against the surface to be wiped.
  Data is collected by Data Collector and the proposed Tactile Transition Network with Parametric Bias (TTNPB) is trained.
  By updating PB online from the current data, the robot can recognize the surface material and perform the wiping operation by controlling the contact according to the desired task definition.
}%
{%
  本研究の全体システムを\figref{figure:system}に示す.
  本研究ではロボットとして低剛性ロボットMyCobotを用いる.
  \figref{figure:system}の右図に示すように, その手先先端に3軸力センサを24個含む接触センサuSkinを搭載し, これを拭きたい表面に押し付けるタスクを考える.
  Data Collectorによりデータを収集し, 提案するTactile Transition Network with Parametric Bias (TTNPB)を訓練する.
  PBを現在のデータからオンライン更新することで表面素材を認識し, 目的となるタスク定義に沿って接触制御を行い拭き動作を実現する.
}%

\subsection{Network Structure} \label{subsec:network-structure}
\switchlanguage%
{%

  The network structure of TTNPB constructed in this study can be expressed as follows:
  \begin{align}
    \bm{F}_{t+1} = \bm{h}_{ttnpb}(\bm{F}_{t}, \bm{x}_{t}, \bm{u}_{t}, \bm{p}) \label{eq:ttnpb}
  \end{align}
  where $t$ is the current time step, $\bm{F}$ is the contact sensor value, $\bm{x}$ is the current hand position, $\bm{u}$ is the control input, $\bm{p}$ is parametric bias (PB), and $\bm{h}_{ttnpb}$ is the prediction model.
  This model represents the transition of contact sensor values when a control input is applied at the current position.
  Since the rattling of a low-rigidity robot body varies depending on its posture, $\bm{x}_{t}$ is added as a network input.
  The contact sensor value $\bm{F}$ provides $xyz$ 3-axis force data for a total of 24 (4$\times$6) locations as shown in the right figure of \figref{figure:system}.
  No force calibration is performed except for the removal of offsets, and the data is handled as dimensionless values.
  $\bm{F}_{\{x, y, z\}}$ denotes the 24-dimensional vector of forces in each direction, and $F^{ave}_{\{x, y, z\}}$ denotes the average of forces in each direction.

  The control input $\bm{u}$ consists of $\{\tau^{ref}_{roll}, \tau^{ref}_{pitch}, f^{ref}_{z}\}$, and the following proportional control changes the actual robot hand position and orientation:
  \begin{align}
    \Delta\theta^{ref}_{roll} &= \max(\Delta\theta^{min}, \min(k_{\theta}(\tau^{ref}_{roll}-\tau_{roll}), \Delta\theta^{max}))\\
    \Delta\theta^{ref}_{pitch} &= \max(\Delta\theta^{min}, \min(k_{\theta}(\tau^{ref}_{pitch}-\tau_{pitch}), \Delta\theta^{max}))\\
    \Delta{x}^{ref}_{z} &= \max(\Delta{x}^{min}_{z}, \min(k_{z}(f^{ref}_{z}-F^{ave}_{z}), \Delta{x}^{max}_{z}))\\
    \theta^{ref}_{roll} &\gets \max(\theta^{min}, \min(\theta^{ref}_{roll} + \Delta\theta^{ref}_{roll}, \theta^{max}))\\
    \theta^{ref}_{pitch} &\gets \max(\theta^{min}, \min(\theta^{ref}_{pitch} + \Delta\theta^{ref}_{pitch}, \theta^{max}))\\
    x^{ref}_{z} &\gets x^{ref}_{z} + \Delta{x}^{ref}_{z}
  \end{align}
  where $\theta^{ref}_{\{roll, pitch, yaw\}}$ is the target hand angle, $x^{ref}_{\{x, y, z\}}$ is the target hand position, $\tau^{ref}_{\{roll, pitch\}}$ is the target torque around the center axis of the contact sensor regarding the roll and pitch axis, $\tau_{\{roll, pitch\}}$ is the current torque around the center axis of the contact sensor, $f^{ref}_{z}$ is the target value of $F^{ave}_{z}$, $k_{\{\theta, z\}}$ is the constant gain of proportional control, and $\{\Delta\theta,\theta, \Delta{x}_{z}\}^{\{min, max\}}$ are the minimum and maximum values of $\{\Delta\theta,\theta, \Delta{x}_{z}\}$.
  The minimum and maximum settings for $x_z$ are excluded because of the possibility of large changes in the $x_z$ direction.
  For $\tau_{\{roll, pitch\}}$, since the distances between the 24 3-axis force sensors are known, it is possible to calculate the current torque using the moment around the center axis of the contact sensors.
  In addition, by setting the target hand angle $\theta^{ref}_{yaw}$ around the yaw axis to 0 and the hand position $x^{ref}_{\{x, y\}}$ in $xy$ direction to arbitrary values, the hand posture $\bm{\theta}^{ref}$ and $\bm{x}^{ref}$ are obtained.
  By solving inverse kinematics on this posture, the target joint angles $\bm{q}^{ref}$ of the robot are obtained and sent to the actual robot.
  Note that even if $\bm{q}^{ref}$ is sent to the robot, it is not possible to realize the exact $\bm{q}^{ref}$ because of the low rigidity of the robot, and a learning mechanism is required.

  Parametric bias $\bm{p}$ is a learnable input variable in neural networks.
  When data with multiple dynamics are collected and trained in a single network, the parameters related to the dynamics that do not fit into the single model self-organize in the space of parametric bias.
  By changing this parameter, the dynamics of $\bm{h}_{ttnpb}$ can be changed to adapt to various surface materials.

  The details of the network configuration are described below.
  In this study, TTNPB consists of 10 layers, which is made up of 4 FC layers (fully-connected layers), 2 LSTM layers (long short-term memory layers) \cite{hochreiter1997lstm}, and 4 FC layers, in order.
  For the number of hidden units, we set \{$N_F+N_x+N_u+N_p$, 300, 200, 100, 100 (number of units in LSTM), 100 (number of units in LSTM), 100, 200, 300, $N_F$\} (note that $N_{\{F, x, u, p\}}$ is the number of dimensions of $\{\bm{F}, \bm{x}, \bm{u}, \bm{p}\}$).
  The activation function is hyperbolic tangent and the update rule is Adam \cite{kingma2015adam}.
  $\bm{p}$ is two-dimensional, and the execution period of \equref{eq:ttnpb} is set to 5 Hz.
  The dimensionality of $\bm{p}$ should be slightly smaller than the expected changes in the body state, because too small a dimensionality will not represent changes in dynamics properly, and too large a dimensionality will make self-organization difficult.
  Also, we set $k_{\{\theta, z\}}=\{0.01, 0.03\}$, $\Delta\theta^{\{min, max\}}=\{-3.0, 3.0\}$ [deg], $\theta^{\{min, max\}}=\{-5.0, 5.0\}$ [deg], and $\Delta{x}_{z}^{\{min, max\}}=\{-5.0, 5.0\}$ [mm].
}%
{%
  本研究で構築するTTNPBのネットワーク構造は以下のように表せる.
  \begin{align}
    \bm{F}_{t+1} = \bm{h}_{ttnpb}(\bm{F}_{t}, \bm{x}_{t}, \bm{u}_{t}, \bm{p}) \label{eq:ttnpb}
  \end{align}
  ここで, $t$は現在のタイムステップ, $\bm{F}$は接触センサ値, $\bm{x}$は現在の手先位置, $\bm{u}$は制御入力, $\bm{p}$はParametric Bias (PB), $\bm{h}_{ttnpb}$は予測モデルを表す.
  本モデルは, 現在の位置において制御入力を加えた際の接触センサ値の遷移を表現している.
  低剛性なロボット身体のガタはその姿勢によっても変化するため, $\bm{x}_{t}$を入力に加えている.
  接触センサ値$\bm{F}$は\figref{figure:system}の右図に示すように4$\times$6の計24箇所についてそれぞれ$xyz$の3軸力データを得ることができる.
  オフセットの除去以外の力の校正はしておらず, 無次元の値として利用する.
  なお, 今後$\bm{F}_{\{x, y, z\}}$はそれぞれの方向における力の24次元ベクトル, $F^{ave}_{\{x, y, z\}}$はそれぞれの方向に関する力の平均を表す.

  制御入力$\bm{u}$は$\{\tau^{ref}_{roll}, \tau^{ref}_{pitch}, f^{ref}_{z}\}$からなり, 以下のP制御により実際のロボットの手先位置姿勢が変化する.
  \begin{align}
    \Delta\theta^{ref}_{roll} &= \max(\Delta\theta^{min}, \min(k_{\theta}(\tau^{ref}_{roll}-\tau_{roll}), \Delta\theta^{max}))\\
    \Delta\theta^{ref}_{pitch} &= \max(\Delta\theta^{min}, \min(k_{\theta}(\tau^{ref}_{pitch}-\tau_{pitch}), \Delta\theta^{max}))\\
    \Delta{x}^{ref}_{z} &= \max(\Delta{x}^{min}_{z}, \min(k_{z}(f^{ref}_{z}-F^{ave}_{z}), \Delta{x}^{max}_{z}))\\
    \theta^{ref}_{roll} &\gets \max(\theta^{min}, \min(\theta^{ref}_{roll} + \Delta\theta^{ref}_{roll}, \theta^{max}))\\
    \theta^{ref}_{pitch} &\gets \max(\theta^{min}, \min(\theta^{ref}_{pitch} + \Delta\theta^{ref}_{pitch}, \theta^{max}))\\
    x^{ref}_{z} &\gets x^{ref}_{z} + \Delta{x}^{ref}_{z}
  \end{align}
  ここで, $\theta^{ref}_{\{roll, pitch, yaw\}}$は指令手先角度, $x^{ref}_{\{x, y, z\}}$は指令手先位置, $\tau^{ref}_{\{roll, pitch\}}$はロール・ピッチ軸に関する接触センサの中心軸周りの指令トルク, $\tau_{\{roll, pitch\}}$は現在の接触センサの中心軸周りのトルク, $f^{ref}_{z}$は$F^{ave}_{z}$の指令値, $k_{\{\theta, z\}}$は比例定数, $\{\Delta\theta,\theta, \Delta{x}_{z}\}^{\{min, max\}}$はそれぞれの値の最小値・最大値を表す.
  $x_z$方向のみ大きく変化する可能性があるため, 最小値・最大値設定を除いている.
  $\tau_{\{roll, pitch\}}$については, 24個の3軸力センサ間の距離がわかっているため, 接触センサの中心軸周りのモーメントから計算可能である.
  これらに加え, yaw軸回りの指令角度$\theta^{ref}_{yaw}$を0, $xy$方向の手先位置$x^{ref}_{\{x, y\}}$を任意に設定することで, 手先の角度$\bm{\theta}^{ref}$と$\bm{x}^{ref}$が得られる.
  これらに対して逆運動学を解くことで, ロボットの指令関節角度$\bm{q}^{ref}$を取得し, 実機へと送る.
  なお, 低剛性なロボットのため$\bm{q}^{ref}$を送っても正確に$\{\bm{\theta}, \bm{x}\}^{ref}$が実現できるわけではなく, 学習機構が必要となる.

  Parametric Bias $\bm{p}$はニューラルネットワークにおける学習可能な入力変数である.
  複数のダイナミクスを持つデータを収集すると, それらを一つのネットワークで学習する際, モデルに入り切らないダイナミクスに関するパラメータが, 入力であるParametric Biasの空間で自己組織化する.
  これを変化させることで, $\bm{h}_{ttnpb}$のダイナミクスを変化させ, 様々な表面素材に対応できるようになる.

  ネットワーク構成の詳細について述べる.
  本研究ではTTNPBは10層としており, 順に4つのFC層(fully-connected layer), 2層のLSTM層(long short-term memory) \cite{hochreiter1997lstm}, 4層のFC層からなる.
  ユニット数については, \{$N_F+N_x+N_u+N_p$, 300, 200, 100, 100 (LSTMのunit数), 100 (LSTMのunit数), 100, 200, 300, $N_F$\}とした(なお, $N_{\{F, x, u, p\}}$は$\{\bm{F}, \bm{x}, \bm{u}, \bm{p}\}$の次元数とする).
  活性化関数はHyperbolic Tangent, 更新則はAdam \cite{kingma2015adam}とした.
  $\bm{p}$は2次元, \equref{eq:ttnpb}の実行周期は5 Hzとしている.
  $\bm{p}$の次元数は小さ過ぎるとdynamicsの変化を適切に表せなくなり, 大きすぎると自己組織化が難しくなるため, 想定される身体状態変化よりも少し小さく設定することが望ましい.
  また, $k_{\{\theta, z\}}=\{0.01, 0.03\}$, $\Delta\theta^{\{min, max\}}=\{-3.0, 3.0\}$ [deg], $\theta^{\{min, max\}}=\{-5.0, 5.0\}$ [deg], $\Delta{x}_{z}^{\{min, max\}}=\{-5.0, 5.0\}$ [mm]としている.
}%

\subsection{Data Collection} \label{subsec:data-collection}
\switchlanguage%
{%

  We describe the data collection method.
  For $\bm{x}$ in the $xy$ direction, the motion is arbitrarily specified, and $\bm{u}$ is changed as follows.
  \begin{align}
    \tau^{ref}_{roll} &\gets \max(\tau^{min}, \min(\tau^{ref}_{roll} + N(0, \sigma_{\theta}), \tau^{max}))\\
    \tau^{ref}_{pitch} &\gets \max(\tau^{min}, \min(\tau^{ref}_{pitch} + N(0, \sigma_{\theta}), \tau^{max}))\\
    f^{ref}_{z} &\gets \max(f^{min}, \min(f^{ref}_{z} + N(0, \sigma_{z}), f^{max}))\\
  \end{align}
  where $N(\mu, \sigma)$ is the Gaussian noise with mean $\mu$ and variance $\sigma$, $\sigma_{\{\theta, z\}}$ is the variance of the data collection for each direction, $\tau^{\{min, max\}}$ are the minimum and maximum values of $\tau$, and $f^{\{min, max\}}$ denotes the minimum and maximum values of $f$.
  By gradually changing $\bm{u}$ within the minimum and maximum values, the target values of the proportional control can be shifted and various data can be obtained.
  In this study, we set $\sigma_{\{\theta, z\}}=\{10, 30\}\;\textrm{or}\;\{3, 10\}$, $\tau^{\{min, max\}}=\{-50, 50\}$, and $f^{\{min, max\}}=\{50, 300\}$.
  As described above, we collect data by changing $\sigma_{\{\theta, z\}}$ to two different values.
  Note that since $F$ is dimensionless, $f$ and $\tau$ are also handled to be dimensionless.
}%
{%
  データ収集の方法について述べる.
  $xy$方向の$\bm{x}$については任意に動きを指定したうえで, $\bm{u}$を以下のように変化させる.
  \begin{align}
    \tau^{ref}_{roll} &\gets \max(\tau^{min}, \min(\tau^{ref}_{roll} + N(0, \sigma_{\theta}), \tau^{max}))\\
    \tau^{ref}_{pitch} &\gets \max(\tau^{min}, \min(\tau^{ref}_{pitch} + N(0, \sigma_{\theta}), \tau^{max}))\\
    f^{ref}_{z} &\gets \max(f^{min}, \min(f^{ref}_{z} + N(0, \sigma_{z}), f^{max}))\\
  \end{align}
  ここで, $N(\mu, \sigma)$は平均$\mu$, 分散$\sigma$のガウス雑音, $\sigma_{\{\theta, z\}}$はそれぞれの方向に対するデータ収集の分散, $\tau^{\{min, max\}}$は$\tau$の最小値と最大値, $f^{\{min, max\}}$は$f$の最小値と最大値を表す.
  最小値と最大値で制限をかけながら$\bm{u}$を徐々に変更していくことで, 比例制御の指令値をずらして様々なデータを取ることができる.
  本研究では$\sigma_{\{\theta, z\}}=\{10, 30\}\;\textrm{or}\;\{3, 10\}$, $\tau^{\{min, max\}}=\{-50, 50\}$, $f^{\{min, max\}}=\{50, 300\}$とする.
  上記のように$\sigma_{\{\theta, z\}}$については2種類に変化させデータを収集する.
  なお, $F$が無次元であるため, $f$と$\tau$も無次元とした.
}%

\subsection{Training of TTNPB}
\switchlanguage%
{%
  The obtained data $D$ is used to train TTNPB.
  By collecting data while changing the surface material, the data can be implicitly embedded in the parametric bias.
  In order to represent each time series transition with different dynamics in a single model, we embed the differences in the dynamics into a low-dimensional space of $\bm{p}$.
  In this case, no labeling of data is required.
  The data collected on the same surface material $k$ is represented as $D_{k}=\{(\bm{F}_{1}, \bm{x}_{1}, \bm{u}_{1}), (\bm{F}_{2}, \bm{x}_{2}, \bm{u}_{2}), \cdots, (\bm{F}_{T_{k}}, \bm{x}_{T_{k}}, \bm{u}_{T_{k}})\}$ ($1 \leq k \leq K$, where $K$ is the total number of trials and $T_{k}$ is the number of steps of the trial for surface material $k$), and the data for learning, $D_{train}=\{(D_{1}, \bm{p}_{1}), (D_{2}, \bm{p}_{2}), \cdots, (D_{K}, \bm{p}_{K})\}$, is obtained.
  Here, $\bm{p}_{K}$ is the parametric bias expressing the dynamics in the surface material $k$, which is a common variable for that surface and another variable for another surface.
  TTNPB is trained using this $D_{train}$.
  In the usual training, only the network weight $W$ is updated, but here $W$ and $p_{k}$ for each state are updated at the same time.
  This means that $p_{k}$ is embedded with the difference in the dynamics of each surface material.
  Mean-squared error is used as the loss function during training, and $\bm{p}_{k}$ is optimized with all initial values set to 0.
}%
{%
  得られたデータ$D$を使いTTNPBを学習させる.
  この際, 表面素材を変化させながらデータを収集することで, これらのデータを暗黙的にParametric Biasに埋め込むことができる.
  異なるダイナミクスを持つそれぞれの時系列データ遷移を一つのモデルで表現できるよう, そのダイナミクスの違いを低次元な$\bm{p}$の空間に埋め込む.
  この際, 一切のラベルづけ等は必要ない.
  ある同一の表面素材$k$において収集されたデータを$D_{k}=\{(\bm{F}_{1}, \bm{x}_{1}, \bm{u}_{1}), (\bm{F}_{2}, \bm{x}_{2}, \bm{u}_{2}), \cdots, (\bm{F}_{T_{k}}, \bm{x}_{T_{k}}, \bm{u}_{T_{k}})\}$ ($1 \leq k \leq K$, $K$は全試行回数, $T_{k}$はその表面素材$k$における試行の動作ステップ数)として, 学習に用いるデータ$D_{train}=\{(D_{1}, \bm{p}_{1}), (D_{2}, \bm{p}_{2}), \cdots, (D_{K}, \bm{p}_{K})\}$を得る.
  ここで, $\bm{p}_{k}$はその表面素材$k$におけるダイナミクスを表現するParametric Biasであり, その状態については共通の変数, 別の状態については別の変数となる.
  この$D_{train}$を使ってTTNPBを学習させる.
  通常の学習ではネットワークの重み$W$のみが更新されるが, ここでは$W$と各状態に関する$p_{k}$が同時に更新される.
  これにより, $p_{k}$にそれぞれの表面素材におけるダイナミクスの違いが埋め込まれることになる.
  学習の際は損失関数として平均二乗誤差を使い, $\bm{p}_{k}$は全て0を初期値として最適化される.
}%

\subsection{Online Recognition of Surface Being Wiped} \label{subsec:online-update}
\switchlanguage%
{%
  Parametric bias $\bm{p}_{k}$ obtained in the previous section is the value at training time, and the current $\bm{p}$ should be appropriate for the current surface material.
  Using the data $D$ obtained for the current state, the current surface material can be estimated by updating the parametric bias online.
  Since only low-dimensional parametric bias $\bm{p}$ is updated, overfitting is unlikely to occur and the system can keep learning.
  This online learning process allows us to obtain a model that is always adapted to the current surface material.

  The number of data obtained is set as $N^{online}_{data}$, and online learning starts when the number of data exceeds $N^{online}_{thre}$.
  Each time new data is received, training is performed with $N^{online}_{batch}$ for the number of batches, $N^{online}_{epoch}$ for the number of epochs, and MomentumSGD for the update rule.
  Data exceeding $N^{online}_{max}$ are deleted from the oldest ones.

  In this study, we set $N^{online}_{\{thre, max\}}=\{10, 30\}$, $N^{online}_{batch}=N^{online}_{max}$, and $N^{online}_{epoch}=3$.
}%
{%
  前節において得られたPB $\bm{p}_{k}$は学習時の値であり, 現在の$\bm{p}$は現在の表面素材に適切な値となるべきである.
  現在の表面素材において得られたデータ$D$を使い, オンラインでParametric Biasを更新することで, 現在の表面素材を推定することができる.
  低次元のParametric Bias $\bm{p}$のみを更新するため過学習は起こりにくく, 常に学習し続けることが可能となる.
  このオンライン学習により, 常に現在の表面素材に適応したモデルを得ることができる.

  得られたデータ数を$N^{online}_{data}$として, データ数が$N^{online}_{thre}$を超えたところからオンライン学習を始める.
  新しいデータが入るたびにバッチ数を$N^{online}_{batch}$, エポック数を$N^{online}_{epoch}$, 更新則をMomentumSGDとして学習を行う.
  $N^{online}_{max}$を超えたデータは古いものから削除していく.

  本研究では, $N^{online}_{\{thre, max\}}=\{10, 30\}$, $N^{online}_{batch}=N^{online}_{max}$, $N^{online}_{epoch}=3$とした.
}%

\subsection{Contact Control with Various Task Definitions} \label{subsec:control}
\switchlanguage%
{%
  We describe the generation of wiping motion using TTNPB.
  We optimize $\bm{u}$ from the loss functions on $\bm{F}$ and $\bm{u}$.
  First, we give the initial value $\bm{u}^{init}_{seq}$ of the time series control input $\bm{u}_{seq}=\bm{u}_{[t:t+N_{step}-1]}$ ($N_{step}$ is the number of TTNPB expansions in the control, or control horizon).
  Let $\bm{u}$ to be optimized be $\bm{u}^{opt}_{seq}$, and repeat the following calculation at time $t$ to obtain the optimal $\bm{u}^{opt}_{t}$.
  \begin{align}
    \bm{F}^{pred}_{seq} &= \bm{h}_{expand}(\bm{F}_{t}, \bm{x}_{t}, \bm{u}^{opt}_{seq})\\
    L &= \bm{h}_{loss}(\bm{F}^{pred}_{seq}, \bm{u}^{opt}_{seq}) \label{eq:control-loss}\\
    \bm{u}^{opt}_{seq} &\gets \bm{u}^{opt}_{seq} - \gamma\partial{L}/\partial{\bm{u}^{opt}_{seq}} \label{eq:control-opt}
  \end{align}
  where $\bm{F}^{pred}_{seq}$ is the predicted $\bm{F}_{[t+1:t+N_{step}]}$, $\bm{h}_{expand}$ is the function $\bm{h}$ expanded $N_{step}$ times, $\bm{h}_{loss}$ is the loss function, and $\gamma$ denotes the learning rate.
  In other words, the future $\bm{F}$ is predicted from the current state $\bm{F}_{t}$ and $\bm{x}_{t}$ by the time series control input $\bm{u}^{opt}_{seq}$, and $\bm{u}^{opt}_{seq}$ is calculated to minimize the loss function set for it using the backpropagation technique and gradient descent method.

  Here, we set $\bm{u}^{init}_{seq}$ as $\bm{u}^{prev}_{\{t+1, \cdots, t+N_{step}-1, t+N_{step}-1\}}$ by using $\bm{u}^{prev}_{[t:t+N_{step}-1]}$ (which is $\bm{u}_{seq}$ optimized in the previous step) shifting it one time step, and duplicating the last term.
  Faster convergence can be obtained by using the previous optimization results.
  For $\gamma$, $[0, \gamma_{max}]$ is exponentially divided into $N^{control}_{batch}$ $\gamma$, and after running \equref{eq:control-opt} on each $\gamma$, $\bm{u}^{opt}_{seq}$ with the smallest loss is selected by computing the loss with \equref{eq:control-loss}, repeating the process $N^{control}_{epoch}$ times.
  Faster convergence can be obtained by always choosing the best learning rate while trying various $\gamma$.

  Here, the loss function $\bm{h}_{loss}$ allows for various task definitions.
  In this study, we set $\bm{h}_{loss}$ to the following three types and conduct experiments with them.
  \begin{align}
    h_{loss, 1}(\bm{F}^{pred}_{seq}) &= ||\bm{F}^{pred}_{z, seq}-\bm{F}^{ref}_{z, seq}||^{2}_{2} \label{eq:loss-1}\\
    h_{loss, 2}(\bm{F}^{pred}_{seq}) &= \textrm{Variance}(\bm{F}^{pred}_{y, seq}) \label{eq:loss-2}\\
    h_{loss, 3}(\bm{F}^{pred}_{seq}) &= ||\bm{F}^{pred}_{right-z, seq}-\bm{F}^{ref}_{seq}||^{2}_{2} + ||\bm{F}^{pred}_{left-z, seq}||^{2}_{2} \label{eq:loss-3}
  \end{align}
  where $\bm{F}^{ref}_{z}$ is the target value of $\bm{F}_{z}$, $||\cdot||_{2}$ is L2 norm, $\textrm{Variance}(\cdot)$ is the time direction average of the variance among sensors at that time, and $\bm{F}^{pred}_{\{right-z, left-z\}}$ is the $\bm{F}^{prev}_{z}$ values of the six rightmost sensors (right) and the rest (left) when the $x$-axis is set as the front.
  In other words, \equref{eq:loss-1} is the error from the target value regarding $\bm{F}_{z}$, \equref{eq:loss-2} is the variance regarding $\bm{F}_{y}$, and \equref{eq:loss-3} is the loss that directs the force to the right regarding $\bm{F}_{z}$.
  In addition, we actually add loss that smoothes the transition of $\bm{u}^{pred}_{seq}$ to each loss.

  In this study, we set $N_{step}=4$, $N^{control}_{batch}=5$, $N^{control}_{epoch}=3$, $\gamma_{max}=0.1$, and $F^{ref}_{z}=200$.
}%
{%
  TTNPBを使った拭き動作生成について述べる.
  ここでは, $\bm{F}$と$\bm{u}$に関する損失関数から, $\bm{u}$を最適化していく.
  まず, 時系列制御入力$\bm{u}_{seq}=\bm{u}_{[t:t+N_{step}-1]}$の初期値$\bm{u}^{init}_{seq}$を与える($N_{step}$は制御におけるTTNPBの展開数, 制御ホライズンを表す).
  最適化する$\bm{u}$を$\bm{u}^{opt}_{seq}$とおき, 時刻$t$において以下の計算を繰り返すことで最適な$\bm{u}^{opt}_{t}$を得る.
  \begin{align}
    \bm{F}^{pred}_{seq} &= \bm{h}_{expand}(\bm{F}_{t}, \bm{x}_{t}, \bm{u}^{opt}_{seq})\\
    L &= \bm{h}_{loss}(\bm{F}^{pred}_{seq}, \bm{u}^{opt}_{seq}) \label{eq:control-loss}\\
    \bm{u}^{opt}_{seq} &\gets \bm{u}^{opt}_{seq} - \gamma\partial{L}/\partial{\bm{u}^{opt}_{seq}} \label{eq:control-opt}
  \end{align}
  ここで, $\bm{F}^{pred}_{seq}$は予測された$\bm{F}_{[t+1:t+N_{step}]}$, $\bm{h}_{expand}$は$\bm{h}$を$N_{step}$回展開した関数, $\bm{h}_{loss}$は損失関数, $\gamma$は学習率を表す.
  つまり, 現在状態$\bm{F}_{t}$と$\bm{x}_{t}$から時系列制御入力$\bm{u}^{opt}_{seq}$により将来の$\bm{F}$を予測し, これに対して設定した損失関数を最小化するように, $\bm{u}^{opt}_{seq}$を誤差逆伝播法と勾配法により最適化する.

  このとき$\bm{u}^{init}_{seq}$は, 前ステップで最適化された$\bm{u}_{seq}$である$\bm{u}^{prev}_{[t:t+N_{step}-1]}$を使い, 時間を一つシフトし最後の項を複製した$\bm{u}^{prev}_{\{t+1, \cdots, t+N_{step}-1, t+N_{step}-1\}}$とする.
  前回の最適化結果を利用することでより速い収束が得られる.
  また, $\gamma$については$[0, \gamma_{max}]$を指数関数的に分割した$N^{control}_{batch}$個の$\gamma$を用意し, それぞれの$\gamma$で\equref{eq:control-opt}を実行した後, \equref{eq:control-loss}で損失を計算し最も損失の小さい$\bm{u}^{opt}_{seq}$を選択することを$N^{control}_{epoch}$回繰り返す.
  様々な$\gamma$を試して最良の学習率を常に選ぶことで, より速い収束が得られる.

  ここで, 損失関数$\bm{h}_{loss}$により様々なタスク定義を行うことができる.
  本研究では$\bm{h}_{loss}$を以下の3種類に設定し比較する.
  \begin{align}
    h_{loss, 1}(\bm{F}^{pred}_{seq}) &= ||\bm{F}^{pred}_{z, seq}-\bm{F}^{ref}_{z, seq}||^{2}_{2} \label{eq:loss-1}\\
    h_{loss, 2}(\bm{F}^{pred}_{seq}) &= \textrm{Variance}(\bm{F}^{pred}_{y, seq}) \label{eq:loss-2}\\
    h_{loss, 3}(\bm{F}^{pred}_{seq}) &= ||\bm{F}^{pred}_{right-z, seq}-\bm{F}^{ref}_{seq}||^{2}_{2} + ||\bm{F}^{pred}_{left-z, seq}||^{2}_{2} \label{eq:loss-3}
  \end{align}
  ここで, $\bm{F}^{ref}_{z}$は$\bm{F}_{z}$の指令値, $\textrm{Variance}(\cdot)$はその時刻のセンサ間の分散に関する時間方向の平均, $\bm{F}^{pred}_{\{right-z, left-z\}}$は$\bm{F}^{pred}_{z}$について, $x$軸を前とした時最も右にあるセンサ6つの値(right)とそれ以外(left)を現している.
  つまり, \equref{eq:loss-1}は$\bm{F}_{z}$に関する指令値との誤差, \equref{eq:loss-2}は$\bm{F}_{y}$に関する分散, \equref{eq:loss-3}は$\bm{F}_{z}$に関して力を右方向に偏らせる損失と言える.
  また, 実際にはそれぞれの損失に対して, $\bm{u}^{pred}_{seq}$の遷移を滑らかにする損失を追加している.

  本研究では, $N_{step}=4$, $N^{control}_{batch}=5$, $N^{control}_{epoch}=3$, $\gamma_{max}=0.1$, $F^{ref}_{z}=200$とする.
}%

\section{Experiments} \label{sec:experiment}

\begin{figure}[t]
  \centering
  \includegraphics[width=0.95\columnwidth]{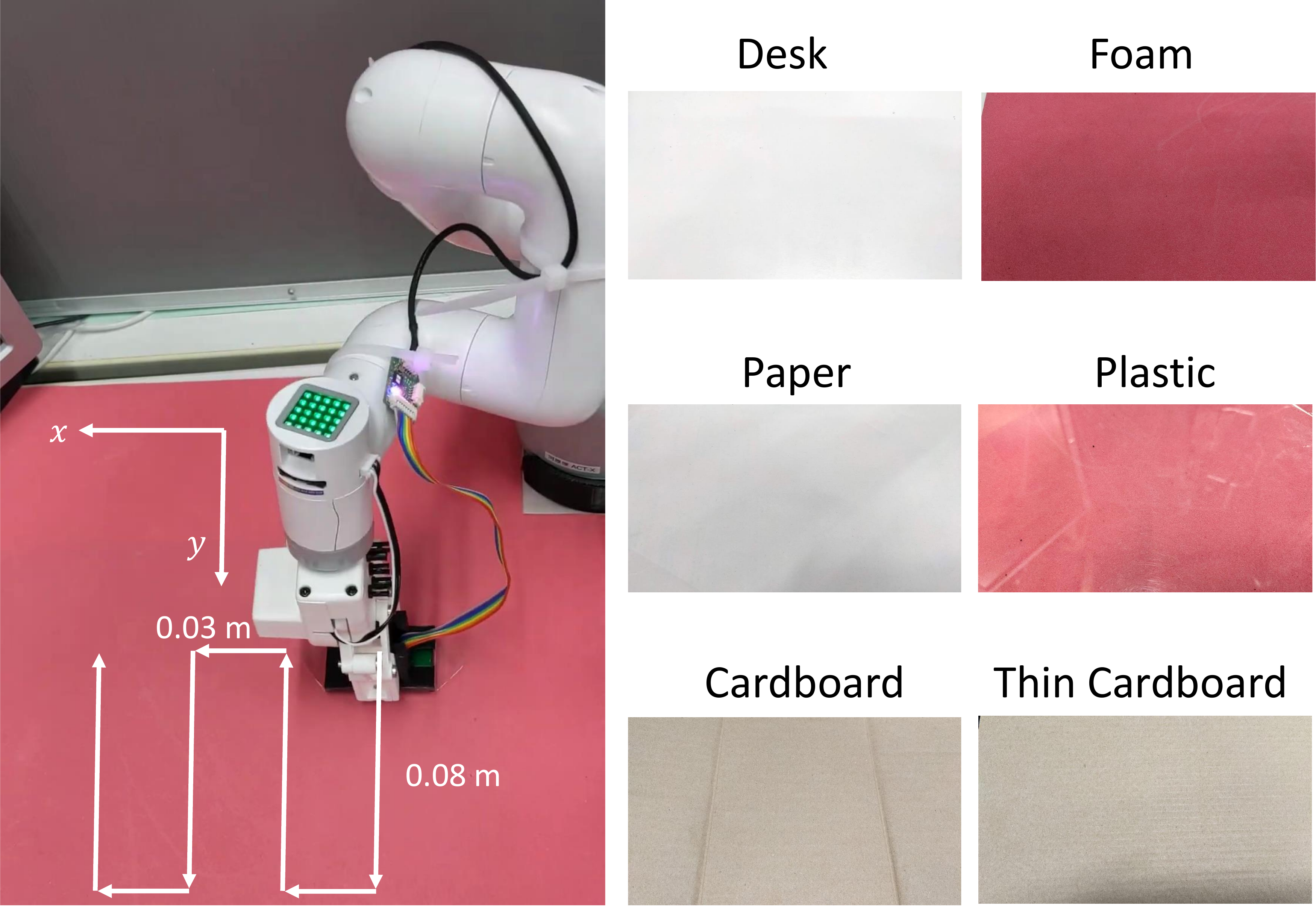}
  \caption{The experimental setup: the trajectory of wiping and various surface materials handled in this study.}
  \label{figure:exp-setup}
  \vspace{-1.0ex}
\end{figure}

\subsection{Experimental Setup}
\switchlanguage%
{%
  In this study, we conduct experiments on the six surface materials shown in the right figure of \figref{figure:exp-setup}.
  Foam is a thin polyethylene foam sheet (2 mm), and Plastic is a thin plastic sheet (0.5 mm).
  Cardboard is thick (5 mm) and wavy, while Thin Cardboard is thin (2 mm) and neither as thick nor as wavy as Cardboard.
  For Paper, Plastic, and Thin Cardboard, the experiments were conducted with these sheets laid on top of Foam.
  The experiments are performed with a slippery tape material wrapped around the uSkin.
  The uSkin moves back and forth along the trajectory in the $xy$ direction as shown in the left figure of \figref{figure:exp-setup}.
}%
{%
  本研究では, \figref{figure:exp-setup}の右図に示す6つの表面素材について実験を行う.
  Foamは薄い発泡性のポリエチレンフォームのシートで, Plasticはプラスチックシートである.
  Cardboardは厚みがあり波打っており, Thin Cardboardは薄く, Cardboardほどの厚みも波打ちも無い.
  Paper, Plastic, Thin Cardboardに関しては, Foamの上にこれらを敷いた状態で実験している.
  uSkinの表面については滑りやすいテープ素材を巻きつけて実験を行う.
  $xy$方向の動作であるが, \figref{figure:exp-setup}の左図のような軌道を行き来する動作を行う.
}%

\begin{figure}[t]
  \centering
  \includegraphics[width=0.95\columnwidth]{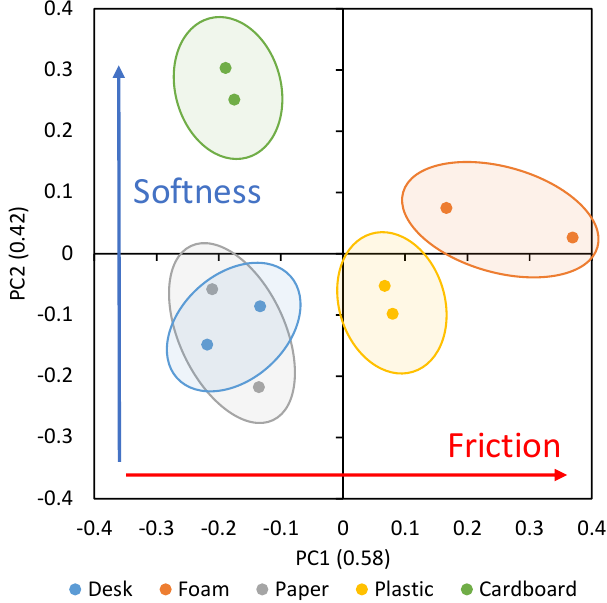}
  \caption{The arrangement of the trained parametric bias.}
  \label{figure:exp-pb}
  \vspace{-1.0ex}
\end{figure}

\begin{figure}[t]
  \centering
  \includegraphics[width=0.95\columnwidth]{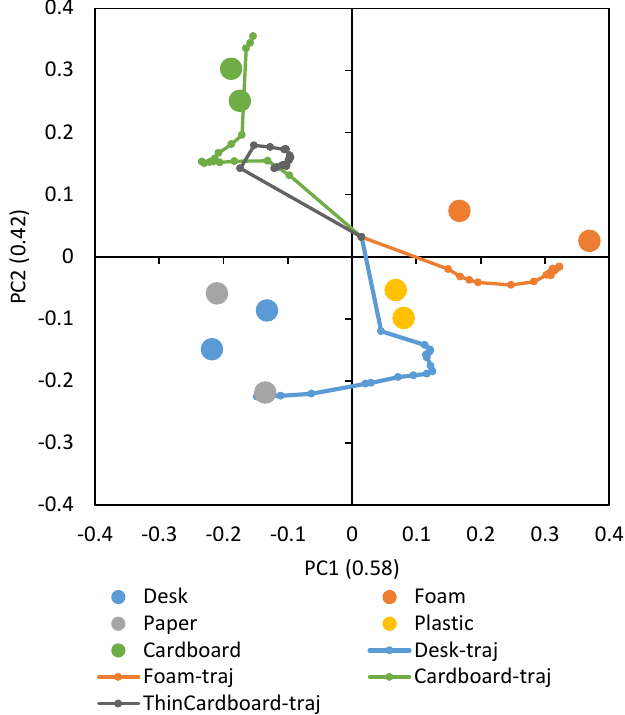}
  \caption{The trajectory of parametric bias while updating it online.}
  \label{figure:exp-pb-traj}
  \vspace{-1.0ex}
\end{figure}

\subsection{Training of TTNPB and Online Learning of Parametric Bias}
\switchlanguage%
{%
  First, we collect data for the five surface materials in \figref{figure:exp-setup}, excluding Thin Cardboard, by the method of \secref{subsec:data-collection}.
  In this case, we change $\sigma_{\{\theta, z\}}$ to two different values as described in \secref{subsec:data-collection}.
  By acquiring data for about 1000 steps in each setting, we use a total of about 10000 data points for training.
  \figref{figure:exp-pb} shows the arrangement of PBs obtained from the training in a 2-dimensional plane by applying Principle Component Analysis (PCA).
  In this study, $\bm{p}$ is set to two dimensions, but even in this case, the principal axes can be obtained by PCA.
  Even when $\sigma$ is set differently, when the surface materials are the same, PBs are placed close to each other because the tactile dynamics are the same.
  It is also considered that PC1 qualitatively represents the axis related to the friction of the surface, and PC2 represents the softness of the surface material.
  Cardboard, Desk, and Paper are very similar in terms of their surface friction, and the friction between each of them and the material of the contact sensor is smaller than that of Plastic and Foam.
  In terms of the softness of the surface materials, Desk, Paper, and Plastic are rigid, and Foam and Cardboard are soft, in that order.
  In other words, even if we do not directly provide the label of the material, it is possible to self-organize the space of PB autonomously depending on whether the dynamics of the material is close or far from another surface material.

  Next, we conducted experiments on the recognition of surface materials.
  We performed the same random motion as in the training phase, with the four surface materials (Desk, Foam, Cardboard, and Thin Cardboard).
  The trajectory of PB when executing \secref{subsec:online-update} is shown in \figref{figure:exp-pb-traj} as ``-traj''.
  It can be seen that the current PBs of Cardboard, Desk, and Foam are each moving toward PBs obtained at the time of training.
  Thin Cardboard is a surface material that was not used in the training, but its trajectory is located slightly below the $\bm{p}_{k}$ of Cardboard.
  This result is consistent with the fact that Thin Cardboard is thinner than Cardboard and thus its surface softness is reduced, while the surface friction does not change much because the material itself is the same.
  For both the surface material used in the training and the unknown surface material, the current dynamics were explored appropriately in the space of PB.
}%
{%
  まず, \figref{figure:exp-setup}におけるThin Cardboardを除いた5つの表面素材について, \secref{subsec:data-collection}の方法によりデータを収集する.
  この際, \secref{subsec:data-collection}で述べたように, $\sigma_{\{\theta, z\}}$を2種類に変化させる.
  それぞれの設定で約1000ステップずつのデータを取得することで, 合計で約10000のデータを用いて学習を行う.
  学習によって得られたPBをPrinciple Component Analysis (PCA)によって2次元平面に配置した図を\figref{figure:exp-pb}に示す.
  本研究では$\bm{p}$は2次元に設定しているが, その場合でもPCAにより主軸を出すことが可能である.
  $\sigma$の設定が違う場合でも, 表面素材が同じ場合は接触覚のダイナミクスが同じであるため近い場所にPBが配置される.
  また, 定性的にPC1が表面の摩擦に関する軸, PC2がその表面素材の柔らかさを表していると考えられる.
  Cardboard, Desk, Paperはその表面摩擦の観点では非常に似通っており, PlasticやFoamに比べると手先の素材との間の摩擦が小さい.
  また, 表面素材の柔らかさの観点では, Desk, Paper, Plasticは剛であり, Foam, Cardboardの順で柔らかい.
  つまり, 直接素材のラベルを与えなくてもそのダイナミクスが近いか遠いかによって自律的なPBの空間の自己組織化が可能である.

  次に, 表面素材の認識について実験を行った.
  表面素材をDesk, Foam, Cardboard, Thin Cardboardの4つに設定した状態で, 学習時と同様のランダム動作を行う.
  \secref{subsec:online-update}を実行した際におけるPBの軌跡を\figref{figure:exp-pb-traj}に``-traj''として示す.
  Cardboard, Desk, Foamについて, それぞれ学習時に得られたPBに近づくように現在のPBが遷移していくことがわかる.
  Thin Cardboardは学習時には用いなかった表面素材であるが, これらはCardboardの$\bm{p}_{k}$よりも少し下の場所に位置した.
  Thin CardboardはCardboardよりも薄いため表面の柔らかさが落ちる一方, 素材自体は同じであるため表面摩擦はあまり変わらないという事実と一致した結果となった.
  学習時に用いた表面素材も, 未知の表面素材も, PBの空間内において適切にダイナミクスを探索することが可能であった.
}%

\begin{figure*}[t]
  \centering
  \includegraphics[width=1.95\columnwidth]{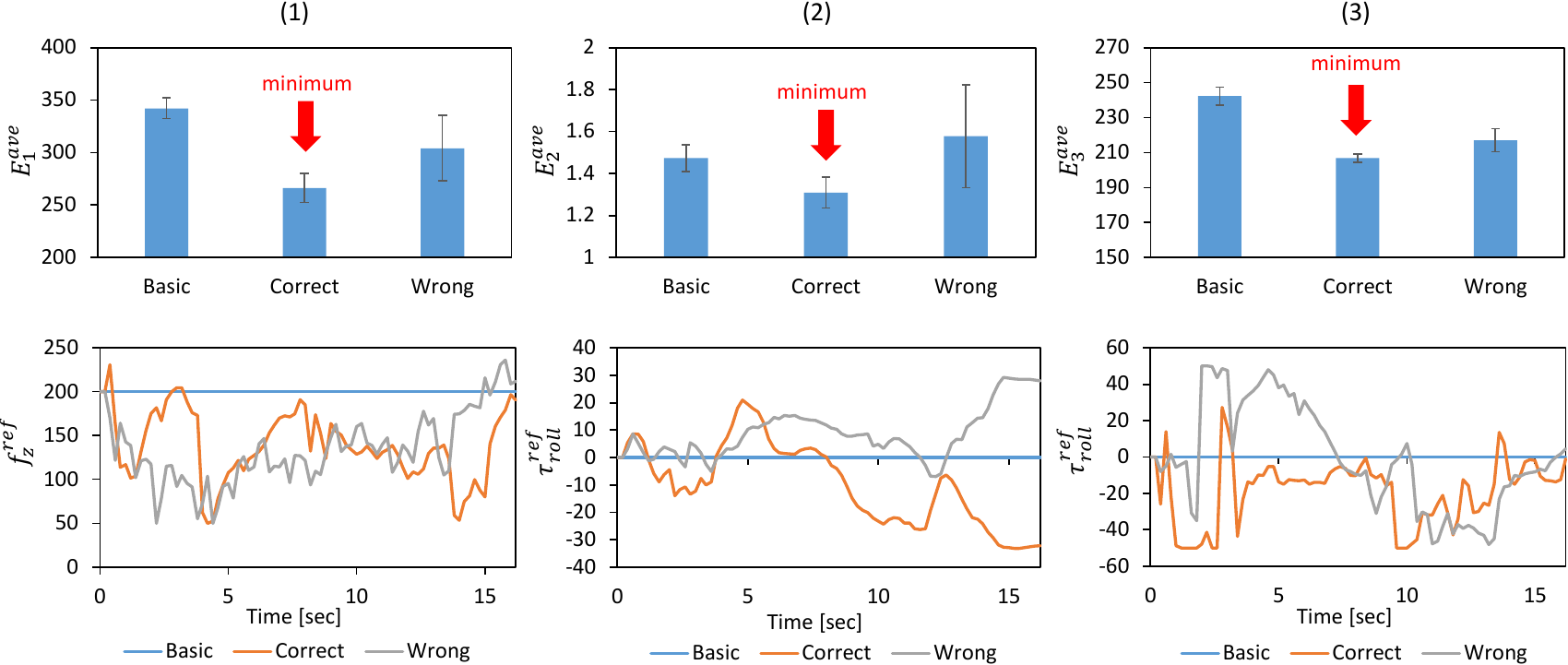}
  \caption{The evaluation value $E^{ave}$ and the transition of the most important control inputs for wiping control with Basic, Correct, and Wrong controllers with loss (1) -- (3).}
  \label{figure:exp-control}
  \vspace{-1.0ex}
\end{figure*}

\subsection{Wiping Behavior Considering Various Surfaces and Task Definitions}
\switchlanguage%
{%
  The three loss functions defined in \secref{subsec:control} are used to execute the wiping operation in each task definition.
  Let Foam be the surface material, Correct be the case where \secref{subsec:control} is executed with the trained PB of Foam, and Wrong be the case where \secref{subsec:control} is executed using PB of a different material, Desk.
  The case where the variance is set to 0 in \secref{subsec:data-collection} is called Basic and is used for comparison.
  The results of the control of Basic, Correct, and Wrong for the three loss functions (the experiments with each loss function are called (1), (2), and (3), respectively) are shown in \figref{figure:exp-control}.
  Here, the control by each loss function is evaluated by the average $E^{ave}$ of the evaluation values $E$ of the same number as shown below:
  \begin{align}
    E_{1} &= ||\bm{F}_{z}-\bm{F}^{ref}_{z}||_{2} \label{eq:eval-1}\\
    E_{2} &= \textrm{Std}(\bm{F}_{y})/F^{abs-ave}_{y} \label{eq:eval-2}\\
    E_{3} &= ||\bm{F}_{right-z}-\bm{F}^{ref}||_{2} + ||\bm{F}_{left-z}||_{2} \label{eq:eval-3}
  \end{align}
  where $\textrm{Std}(\cdot)$ denotes the standard deviation, and $F^{abs-ave}_{y}$ denotes the mean of the absolute values of $\bm{F}_{y}$.
  For $E_{2}$, since the standard deviation increases as the absolute value of $F_{y}$ increases, the value is evaluated by dividing it by $F^{abs-ave}_{y}$.
  The smaller this evaluation value is, the better.
  We also show the transition of $F^{ref}$ for (1) and $\tau_{roll}$ for (2) and (3) as the most important values of $\bm{u}$ in the control for each loss function.

  For each loss, the Correct case generates the best controller with the smallest $E^{ave}$.
  For (1), the generated $F^{ref}_{z}$ transitions show that both Correct and Wrong output smaller values than $F^{ref}_{z}$, which is constant in Basic.
  This may be due to the fact that in Basic, where $F^{ref}_{z}=200$ is constant, the tips of the hand are often caught by the surface materials and a large $\bm{F}_{z}$ is generated, causing a large increase in \equref{eq:loss-1}.
  Therefore, a value smaller than $F^{ref}_{z}=200$ is properly output in alignment with the current $\bm{x}$.
  As for (2), the transition of the generated $\tau^{ref}_{roll}$ shows that $\tau^{ref}_{roll}$ repeatedly increases and decreases along with the motion.
  Since the direction of motion changes about the $y$-axis as in \figref{figure:exp-setup} in this study, the roll angle is changed accordingly to avoid a situation where the sensor is trapped by the surface material or a strong force is applied to some of the sensors.
  As for (3), the generated transition of $\tau^{ref}_{roll}$ shows that it swings in the negative direction, especially for Correct.
  Since the purpose of this experiment is to move the contact sensor by applying force to the right side of the sensor and not to the left side, the torque in the roll direction that is applied to the contact sensor becomes as negative as possible.
}%
{%
  \secref{subsec:control}で定義した3つの損失関数を使い, それぞれのタスク定義において拭き動作を実行する.
  表面素材をFoamとし, 学習された際のFoamのPBによって\secref{subsec:control}を実行する場合をCorrect, 異なるDeskのPBを用いて制御する場合をWrongとする.
  また, \secref{subsec:data-collection}において分散を0とした動作をBasicと呼び, 比較対象とする.
  3つの損失関数において(それぞれの損失を使った実験を(1), (2), (3)と呼ぶ), Basic, Correct, Wrongの制御を行った際の結果を\figref{figure:exp-control}に示す.
  ここで, それぞれの損失関数による制御は同番号の評価値$E$の平均$E^{ave}$によって評価する.
  \begin{align}
    E_{1} &= ||\bm{F}_{z}-\bm{F}^{ref}_{z}||_{2} \label{eq:eval-1}\\
    E_{2} &= \textrm{Std}(\bm{F}_{y})/F^{abs-ave}_{y} \label{eq:eval-2}\\
    E_{3} &= ||\bm{F}_{right-z}-\bm{F}^{ref}||_{2} + ||\bm{F}_{left-z}||_{2} \label{eq:eval-3}
  \end{align}
  ここで, $\textrm{Std}(\cdot)$は標準偏差, $F^{abs-ave}_{y}$は$\bm{F}_{y}$の絶対値の平均を表す.
  $E_{2}$については, $F_{y}$の絶対値が大きいほど標準偏差も上昇するため, $F^{abs-ave}_{y}$で割ることで評価値としている.
  この評価値は小さければ小さいほど良い値である.
  また, それぞれの損失関数で制御する際の$\bm{u}$の中でも, 最も重要と考えられる値として, (1)については$F^{ref}$, (2)と(3)については$\tau_{roll}$の遷移を表示する.

  それぞれの損失について, Correctの場合が最も$E$が小さく, 良いコントローラが生成されている.
  (1)については, 生成された$f^{ref}_{z}$の遷移を見ると, Correct, WrongともにBasicでは一定である$F^{ref}_{z}$よりも小さな値が出力されている.
  これは, $F^{ref}_{z}=200$で一定のBasicでは良く手先が表面素材に引っかかり, 大きな$\bm{F}_{z}$が生成され, \equref{eq:loss-1}が大きく上昇してしまうことが要因と考えられる.
  そのため, $F^{ref}_{z}=200$よりも小さな値を現在の$\bm{x}$に沿って適切に出力している.
  (2)については, 生成された$\tau^{ref}_{roll}$の遷移を見ると, 動作に沿って$\tau^{ref}_{roll}$が増加と減少を繰り返していることがわかる.
  本研究では\figref{figure:exp-setup}のように$y$軸について移動方向が変化するため, それに応じて傾ける角度を変え, 表面素材へ引っかかったり, 一部のセンサに強い力がかかってしまう状態を避けていると考えられる.
  (3)について, 生成された$\tau^{ref}_{roll}$の遷移を見ると, 特にCorrectについてはマイナス方向に振れている.
  本実験は接触センサの右側に力を加え, 左側に力を加えないように動かす動作であるため, 接触センサにかかるroll方向のトルクをなるべくマイナスした動作が生成されている.
}%

\section{Discussion} \label{sec:discussion}
\switchlanguage%
{%
  We discuss the experimental results in this study.
  First, parametric bias was able to self-organize the dynamics of surface materials into the space of PB based on the differences in dynamics among the data, even without providing explicit labels.
  The dynamics are automatically classified by human intuitive axes such as surface friction and surface softness.
  By updating only the parametric bias, it was possible to recognize the surface material currently being wiped from the current data.
  It was also shown that the model can be applied to unknown surface materials.
  Next, it was found that the model can be used to represent various tasks by loss functions.
  The loss function can be freely taken for the contact sensor values, and the task representation can be changed.
  Our method outperforms the default controller with constant control input.
  On the other hand, the performance may suffer if the current recognition of the surface material is not accurate, and so it is necessary to simultaneously recognize the surface material and control the contact.

  The limitations of this study are as follows.
  In this study, the experiment was conducted by first establishing proportional control, and then changing the target value of the proportional control, which resulted in stable operation.
  Although it is possible to use angular velocity and other parameters directly as control inputs without proportional control, it caused the behavior to become very unstable.
  The use of more complicated control inputs is a future task to solve.
  In addition, this study deals only with two-dimensional planes.
  Of course, there is a possibility that 3D surfaces can also be handled, but since feedback alone is not sufficient for this, it is necessary to combine this method with trajectory generation based on shape recognition using images and point clouds.
  This is also a challenge for the future.
}%
{%
  本研究の実験について考察する.
  まず, Parametric Biasは明確なラベルを与えずとも, そのデータ間におけるダイナミクスの違いから, 表面素材のダイナミクスをPBの空間に自己組織化することが可能であった.
  これは, 表面摩擦や表面の柔らかさ等, 人間に直感的な軸によって分類されていた.
  そして, Parametric Biasのみを更新することで, 現在のデータから現在拭いている表面素材を認識することが可能であった.
  そしてこれは, 未知の表面素材についても適用可能であることが示された.
  次に, 本モデルを用いることで, 様々なタスクを損失関数によって表現し, これを実現することが可能であるとわかった.
  接触センサの値について自由に損失関数を取り, タスク表現を変更できる.
  制御入力一定のデフォルトのコントローラに対して, 本手法はその性能を上回る.
  一方, 現在の表面素材の認識が正確でない場合は性能が落ちることがあり, 表面素材認識と制御は同時に行っていく必要がある.

  本研究の限界について述べる.
  本研究は, 最初に比例制御を組み, その上でその目標値を変更することで実験を行ったため, 動作が安定していた.
  P制御を組まずに角速度等を直接制御入力に使うことも可能であるが, 動作が非常に不安定になってしまったため難しかった.
  より複雑な制御入力に利用は今後の課題である.
  また, 本研究は2次元平面のみを扱っている.
  もちろん3次元曲面も対応できる可能性があるが, これはフィードバックだけでは間に合わないため, 画像や点群を使った形状認識からの軌道生成と合わせる必要があり, 今後の課題である.
}%

\section{CONCLUSION} \label{sec:conclusion}
\switchlanguage%
{%
  In this study, we have constructed a neural network representing state transitions of contact sensors in order to realize a wiping motion applicable to various surface materials and task definitions.
  The network is trained by collecting data from the motion with random noise added to the proportional control.
  By using parametric bias as a network input, information on the characteristics of the surface material can be embedded in the network.
  The current properties of the surface material can be estimated from the contact state transitions, and the motion can be changed accordingly.
  By changing the loss function related to the contact state according to the task, it is possible to apply force evenly or to only certain areas.
  In addition, we have shown through experiments that these learning behaviors can be applied to a low-rigidity robot that has difficulty in making precise movements.
  As this study dealt only with two-dimensional planes, we would like to apply this method to three-dimensional curved surfaces to make it possible to execute motions in a more practical manner.
}%
{%
  本研究では, 様々な表面素材とタスク目的に対応可能な拭き動作の実現を目指し, 接触センサの状態遷移を表現したネットワークを構築した.
  比例制御にランダムノイズを加えた動作からデータを収集し, これを訓練する.
  入力としてParametric Biasを用いることで, 表面素材の特性に関する情報をネットワークに埋め込むことができる.
  接触状態遷移からの現在の表面素材特性を推定し, それに応じて動作を変化させることができる.
  タスクに応じて接触状態に関する損失関数を変化させることで, 満遍なく力をかけたり, 一部の角に力をかけたような動きが可能となる.
  また, これらの学習動作が, 正確な動作の難しい低剛性なロボットに適用できることを, 実機実験により示した.
  本研究では2次元平面のみを扱ったが, 本手法を3次元曲面に適用し, より実用的な形で動作実行を可能としていきたい.
}%

{
  %\footnotesize
  %\small
  %\bibliographystyle{junsrt}
  \bibliographystyle{IEEEtran}
  \bibliography{main}

\begin{thebibliography}{10}
\providecommand{\url}[1]{#1}
\csname url@rmstyle\endcsname
\providecommand{\newblock}{\relax}
\providecommand{\bibinfo}[2]{#2}
\providecommand\BIBentrySTDinterwordspacing{\spaceskip=0pt\relax}
\providecommand\BIBentryALTinterwordstretchfactor{4}
\providecommand\BIBentryALTinterwordspacing{\spaceskip=\fontdimen2\font plus
\BIBentryALTinterwordstretchfactor\fontdimen3\font minus
  \fontdimen4\font\relax}
\providecommand\BIBforeignlanguage[2]{{%
\expandafter\ifx\csname l@#1\endcsname\relax
\typeout{** WARNING: IEEEtran.bst: No hyphenation pattern has been}%
\typeout{** loaded for the language `#1'. Using the pattern for}%
\typeout{** the default language instead.}%
\else
\language=\csname l@#1\endcsname
\fi
#2}}

\bibitem{kim2019cleaning}
J.~Kim, A.~K. Mishra, R.~Limosani, M.~Scafuro, N.~Cauli, J.~Santos-Victor,
  B.~Mazzolai, and F.~Cavallo, ``{Control strategies for cleaning robots in
  domestic applications: A comprehensive review},'' \emph{International Journal
  of Advanced Robotic Systems}, vol.~16, no.~4, pp. 1--21, 2019.

\bibitem{hess2012coverage}
J.~Hess, G.~D. Tipaldi, and W.~Burgard, ``{Null space optimization for
  effective coverage of 3D surfaces using redundant manipulators},'' in
  \emph{Proceedings of the 2012 IEEE/RSJ International Conference on
  Intelligent Robots and Systems}, 2012, pp. 1923--1928.

\bibitem{domentios2018wiping}
A.~C. Dometios, Y.~Zhou, X.~S. Papageorgiou, C.~S. Tzafestas, and T.~Asfour,
  ``{Vision-Based Online Adaptation of Motion Primitives to Dynamic Surfaces:
  Application to an Interactive Robotic Wiping Task},'' \emph{IEEE Robotics and
  Automation Letters}, vol.~3, no.~3, pp. 1410--1417, 2018.

\bibitem{saito2020wiping}
N.~Saito, D.~Wang, T.~Ogata, H.~Mori, and S.~Sugano, ``{Wiping 3D-objects using
  Deep Learning Model based on Image/Force/Joint Information},'' in
  \emph{Proceedings of the 2020 IEEE/RSJ International Conference on
  Intelligent Robots and Systems}, 2020, pp. 10\,152--10\,157.

\bibitem{saito2022wiping}
N.~Saito, T.~Shimizu, T.~Ogata, and S.~Sugano, ``{Utilization of
  Image/Force/Tactile Sensor Data for Object-Shape-Oriented Manipulation:
  Wiping Objects With Turning Back Motions and Occlusion},'' \emph{IEEE
  Robotics and Automation Letters}, vol.~7, no.~2, pp. 968--975, 2022.

\bibitem{martin2019variable}
R.~Mart\'in-Mart\'in, M.~A. Lee, R.~Gardner, S.~Savarese, J.~Bohg, and A.~Garg,
  ``{Variable Impedance Control in End-Effector Space: An Action Space for
  Reinforcement Learning in Contact-Rich Tasks},'' in \emph{Proceedings of the
  2019 IEEE/RSJ International Conference on Intelligent Robots and Systems},
  2019, pp. 1010--1017.

\bibitem{wang2021variable}
C.~Wang, Z.~Kuang, X.~Zhang, and M.~Tomizuka, ``{Safe Online Gain Optimization
  for Variable Impedance Control},'' arXiv preprint arXiv:2111.01258, 2019.

\bibitem{tani2002parametric}
J.~Tani, ``{Self-organization of behavioral primitives as multiple attractor
  dynamics: a robot experiment},'' in \emph{Proceedings of the 2002
  International Joint Conference on Neural Networks}, 2002, pp. 489--494.

\bibitem{tani2004parametric}
J.~Tani, M.~Ito, and Y.~Sugita, ``{Self-organization of distributedly
  represented multiple behavior schemata in a mirror system: reviews of robot
  experiments using RNNPB},'' \emph{Neural Networks}, vol.~17, no.~8, pp.
  1273--1289, 2004.

\bibitem{kawaharazuka2020dynamics}
K.~Kawaharazuka, K.~Tsuzuki, M.~Onitsuka, Y.~Asano, K.~Okada, K.~Kawasaki, and
  M.~Inaba, ``{Object Recognition, Dynamic Contact Simulation, Detection, and
  Control of the Flexible Musculoskeletal Hand Using a Recurrent Neural Network
  With Parametric Bias},'' \emph{IEEE Robotics and Automation Letters}, vol.~5,
  no.~3, pp. 4580--4587, 2020.

\bibitem{kawaharazuka2022cloth}
K.~Kawaharazuka, A.~Miki, M.~Bando, K.~Okada, and M.~Inaba, ``{Dynamic Cloth
  Manipulation Considering Variable Stiffness and Material Change Using Deep
  Predictive Model With Parametric Bias},'' \emph{Frontiers in Neurorobotics},
  vol.~16, pp. 1--16, 2022.

\bibitem{kawaharazuka2022vservoing}
K.~Kawaharazuka, N.~Kanazawa, K.~Okada, and M.~Inaba, ``{Self-Supervised
  Learning of Visual Servoing for Low-Rigidity Robots Considering Temporal Body
  Changes},'' \emph{IEEE Robotics and Automation Letters}, vol.~7, no.~3, pp.
  7881--7887, 2022.

\bibitem{tomo2018uskin}
T.~P. Tomo, A.~Schmitz, W.~K. Wong, H.~Kristanto, S.~Somlor, J.~Hwang,
  L.~Jamone, and S.~Sugano, ``{Covering a Robot Fingertip With uSkin: A Soft
  Electronic Skin With Distributed 3-Axis Force Sensitive Elements for Robot
  Hands},'' \emph{IEEE Robotics and Automation Letters}, vol.~3, no.~1, pp.
  124--131, 2018.

\bibitem{hochreiter1997lstm}
S.~Hochreiter and J.~Schmidhuber, ``{Long short-term memory},'' \emph{Neural
  computation}, vol.~9, no.~8, pp. 1735--1780, 1997.

\bibitem{kingma2015adam}
D.~P. Kingma and J.~Ba, ``{Adam: A Method for Stochastic Optimization},'' in
  \emph{Proceedings of the 3rd International Conference on Learning
  Representations}, 2015, pp. 1--15.

\end{thebibliography}
}

\end{document}